\documentclass[10pt,twocolumn,letterpaper]{article}

\usepackage{iccv}
\usepackage{times}
\usepackage{epsfig}
\usepackage{graphicx}
\usepackage{diagbox} 
\usepackage{amsmath}
\usepackage{amssymb}
\usepackage{algorithm}
\usepackage{booktabs}
\usepackage{lipsum}
\usepackage{algorithm}  
\usepackage{algpseudocode}  
\usepackage{amsmath}
\usepackage{multirow}
\usepackage[dvipsnames,svgnames,x11names]{xcolor}
\usepackage{enumitem}
\setitemize{noitemsep,topsep=0pt,parsep=0pt,partopsep=0pt}
\usepackage[accsupp]{axessibility} 

\newcommand{\ptq}[0]{\texttt{P}\texttt{\small TQ4DM}}

\usepackage[breaklinks=true,bookmarks=false]{hyperref}
\hypersetup{
    colorlinks=true,
    citecolor=Teal,
}

\iccvfinalcopy 

\def\iccvPaperID{8235} 

\def\eg{\emph{e.g.}{}} 
\def\ie{\emph{i.e.}}

\def\wrt{w.r.t.} 
\def\etal{\emph{et al.}}

\newcommand*\samethanks[1][\value{footnote}]{\footnotemark[#1]}

\ificcvfinal\fi

\begin{document}

\title{Post-training Quantization on Diffusion Models}
\author{
  \textbf{Yuzhang Shang$^{1,4}$\thanks{\, Equal contribution. $\dag$\,  Corresponding author.}, Zhihang Yuan$^2$\samethanks{}, Bin Xie$^{1}$, Bingzhe Wu$^{3}$,}
  \textbf{Yan Yan$^{1\dag}$}\\
  $^{1}$Illinois Institute of Technology,
  $^{2}$Houmo AI, 
  $^{3}$Tencent AI Lab,
  $^{4}$Cisco Research \vspace{-3pt}\\
  \tt\small{\{yshang4, bxie9\}@hawk.iit.edu, zhihang.yuan@huomo.ai} \vspace{-3pt}\\
  \tt\small{yuzshang@cisco.com, bingzhewu@tencent.com, yyan34@iit.edu} 
}



\maketitle

\ificcvfinal\thispagestyle{empty}\fi


\begin{abstract}
Denoising diffusion (score-based) generative models have recently achieved significant accomplishments in generating realistic and diverse data. 
Unfortunately, the generation process of current denoising diffusion models is notoriously slow due to the lengthy iterative noise estimations, which rely on cumbersome neural networks. It prevents the diffusion models from being widely deployed, especially on edge devices. 
Previous works accelerate the generation process of diffusion model (DM) via finding shorter yet effective sampling trajectories. However, they overlook the cost of noise estimation with a heavy network in every iteration.
In this work, we accelerate generation from the perspective of compressing the noise estimation network. 
Due to the difficulty of retraining DMs, we exclude mainstream training-aware compression paradigms and introduce post-training quantization (PTQ) into DM acceleration. 
However, the output distributions of noise estimation networks change with time-step, making previous PTQ methods fail in DMs since they are designed for single-time step scenarios.
To devise a DM-specific PTQ method, we explore PTQ on DM in three aspects: quantized operations, calibration dataset, and calibration metric. 
We summarize and use several observations derived from all-inclusive investigations to formulate our method, which especially targets the unique multi-time-step structure of DMs. 
Experimentally, our method can directly quantize full-precision DMs into 8-bit models while maintaining or even improving their performance in a training-free manner. 
Importantly, our method can serve as a plug-and-play module on other fast-sampling methods, e.g., DDIM~\cite{nichol2021improved}. 
The code is available at \url{https://github.com/42Shawn/PTQ4DM}.
\end{abstract}

\section{Introduction}
\label{sec:intro}

Recently, denoising diffusion (also dubbed score-based) generative models~\cite{ho2020denoising,song2019generative,song2019generative,song2020score} have achieved phenomenal success in various generative tasks, such as images~\cite{ho2020denoising,song2020score,nichol2021improved}, audio~\cite{mittal2021symbolic}, video~\cite{singer2022make}, and graphs~\cite{niu2020permutation}. Besides these fundamental tasks, their flexibility of implementation on downstream tasks is also attractive, \eg, they are effectively introduced for super-resolution~\cite{saharia2022image,kadkhodaie2020solving}, inpainting~\cite{song2020score,kadkhodaie2020solving}, and image-to-image translation~\cite{sasaki2021unit}. Diffusion models (DMs) have achieved superior performances on most of these tasks and applications, both concerning quality and diversity, compared with historically SoTA Generative Adversarial Networks (GANs)~\cite{goodfellow2020generative}.    

A diffusion process transforms real data gradually into Gaussian noise, and then the process is reversed to generate real data from Gaussian noise (denoising process)~\cite{ho2020denoising,yang2022diffusionsurvey}. 
Particularly, the denoising process requires iterating the noise estimation (also known as a score function~\cite{song2020score}) via a cumbersome neural network over thousands of time-steps. 
While it has a compelling quantity of images, its long iterative process and high inference cost for generating samples make it undesirable. 
Thus, increasing the speed of this generation process is now an active area of research~\cite{chen2020wavegrad,san2021noise,nichol2021improved,song2020score,bao2022analytic,lu2022dpm}.
To accelerate diffusion models, researchers propose several approaches, which mainly focus on sample trajectory learning for faster sampling strategies. For example, Chen \etal~\cite{chen2020wavegrad} and San-Roman \etal~\cite{san2021noise} propose faster step size schedules for VP diffusions that still yield relatively good quality/diversity metrics; Song \etal~\cite{song2020denoising} adopt implicit phases in the denoising process; Bao \etal~\cite{bao2022analytic} and Lu \etal~\cite{lu2022dpm} derive analytical approximations to simplify the generation process. 

\sloppy
Our study suggests that two orthogonal factors slow down the denoising process: i) lengthy iterations for sampling images from noise, and ii) a cumbersome network for estimating noise in each iteration. 
Previously DM acceleration methods only focus on the former~\cite{chen2020wavegrad,san2021noise,nichol2021improved,song2020score,bao2022analytic,lu2022dpm}, but overlook the latter. 
From the perspective of network compression, many popular network quantization and pruning methods follow a simple pipeline: training the original model and then fine-tuning the quantized/pruned compressed model~\cite{li2021brecq,shang2021lipschitz}. Particularly, this training-aware compression pipeline requires a full training dataset and many computation resources to perform end-to-end backpropagation. 
For DMs, however, 1) training data are not always ready-to-use due to privacy and commercial concerns; 2) the training process is extremely expensive. 
For example, there is no access to the training data for the industry-developed text-to-image models Dall·E2~\cite{ramesh2022hierarchicaldalle2} and Imagen~\cite{saharia2022photorealisticimagen}. Even if one can access their datasets, fine-tuning them also consumes hundreds of thousands of GPU hours. Those two obstacles make training-aware compression not suitable for DMs. 

Training-free network compression techniques are what we need for DM acceleration. Therefore, we propose to introduce post-training quantization (PTQ)~\cite{nagel2021whitepaper,cai2020zeroq,li2021brecq} into DM acceleration. In a training-free manner, PTQ can not only speed up the computation of the denoising process but also reduce the resources to store the diffusion model weight, which is required in DM acceleration. 
Although PTQ has many attractive benefits, its implementation in DMs remains challenging. 
The main reason is that the structure of DMs is hugely different from previously PTQ-implemented structures (\eg, CNN and ViT for image recognition). Specifically, the output distributions of noise estimation networks change with time-step, making previous PTQ methods fail in DMs since they are designed for single-time-step scenarios.

\begin{figure}
    \centering
    \includegraphics[width=0.4\textwidth]{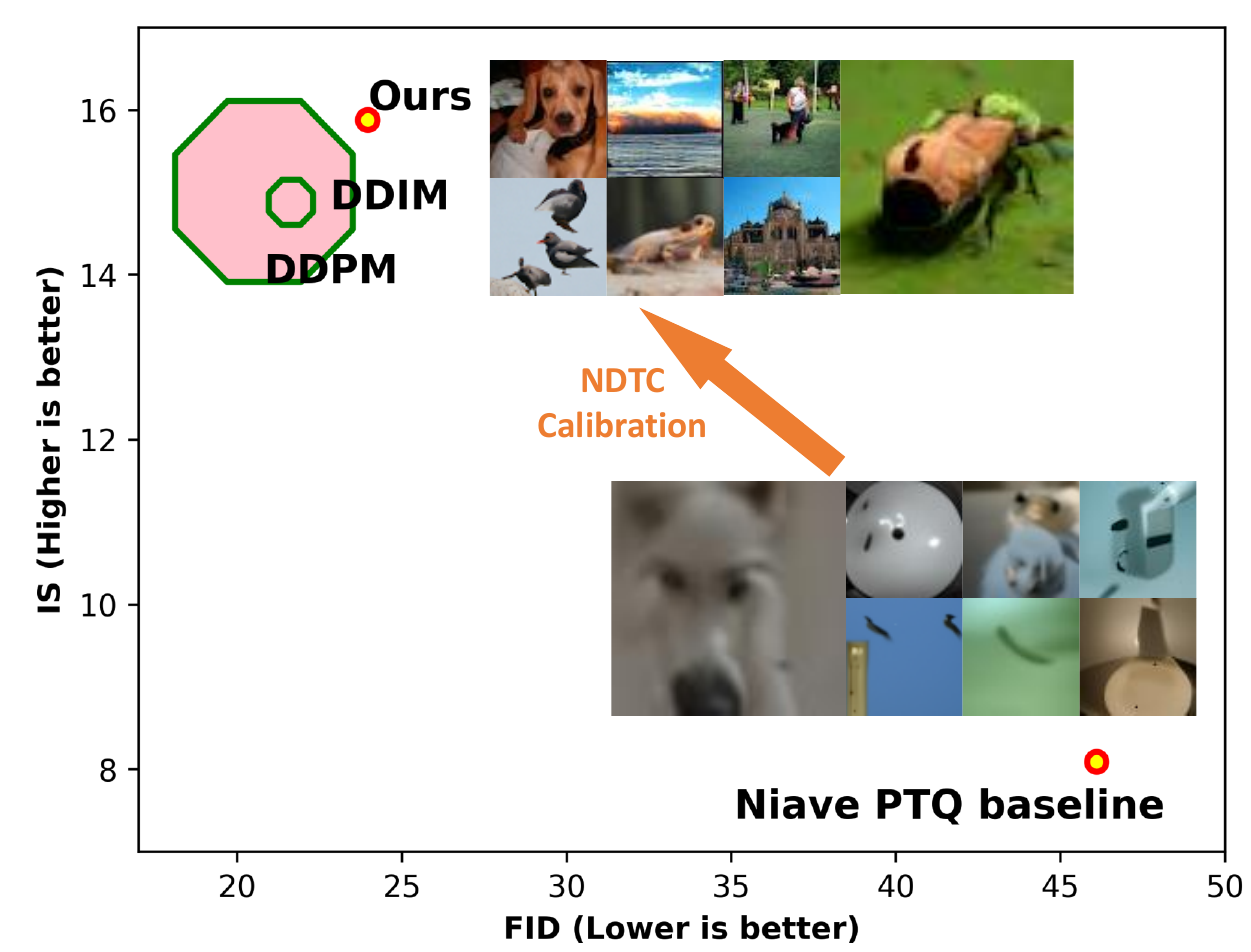}
    \caption{Performance summary on ImageNet64. The X-axis and Y-axis denote the performance \wrt FID score and Inception Score, respectively. Note that the center (not the boundary) of the dot corresponds to the model performance. The \underline{size of the dots} denotes theoretical inference time.}
    \label{fig:performance_summary}
    \vspace{-0.2in}
\end{figure}

This study attempts to answer the following fundamental question: How does the design of the core ingredients of the PTQ for the DMs process (\eg, quantized operation selection, calibration set collection, and calibration metric) affect the final performance of the quantized diffusion models? 
To this end, we analyze the PTQ and DMs individually and correlatedly. 
We find that simple generalizations of previous PTQ methods to DMs lead to huge performance drops due to output distribution discrepancies \wrt time-step in the denoising process. In other words, noise estimation networks rely on time-step, which makes their output distributions change with time-step. This means that a key module of the previous PTQ calibration, cannot be used in our case. 
Based on the above observations, we devise a DM-specific calibration method, termed Normally Distributed Time-step Calibration (\emph{NDTC}), which first samples a set of time-steps from a skew normal distribution, and then generates calibration samples in terms of sampled time-steps by the denoising process. In this way, the time-step discrepancy in the calibration set is enhanced, which improves the performance of~\ptq. Finally, we propose a novel DM acceleration method, Post-Training for Diffusion Models (\ptq) via incorporating all the explorations. 

Overall, the contributions of this paper are three-fold:
    \textbf{(i)} To accelerate denoising diffusion models, we introduce PTQ into DM acceleration where noise estimation networks are directly quantized in a post-training manner. To the best of our knowledge, this is the first work to investigate diffusion model acceleration from the perspective of training-free network compression.  
    \textbf{(ii)} After all-inclusively investigations of PTQ and DMs, we observe the performance drop induced by PTQ for DMs can be attributed to the discrepancy of output distributions in various time-steps. Targeting this observation, we explore PTQ from different aspects and propose~\ptq.
    \textbf{(iii)} Experimentally,~\ptq~can quantize the pre-trained diffusion models to 8-bit without significant performance loss for the first time. Importantly, \ptq~can serve as a plug-and-play module for other SoTA DM acceleration methods, as shown in Fig.~\ref{fig:performance_summary}.

\section{Related Work}
\label{related}

\subsection{Diffusion Model Acceleration}
Due to the long iterative process in conjunction with the high cost of denoising via networks, diffusion models cannot be widely implemented. To accelerate the diffusion probabilistic models (DMs), previous works pursue finding shorter sampling trajectories while maintaining the DM performance. Chen~\etal~\cite{chen2020wavegrad} introduce grid search and find an effective trajectory with only six time-steps. However, the grid search approach can not be generalized into very long trajectories subject to its exponentially growing time complexity. Watson~\etal~\cite{watson2021learning} model the trajectory searching as a dynamic programming problem. Song~\etal~\cite{song2020improved} construct a class of non-Markovian diffusion processes that lead to the same training objective, but whose reverse process can be much faster to sample from. As for DMs with continuous timesteps (\ie, score-based perspective~\cite{song2020score}), Song \etal~\cite{song2019generative,song2020score} formulate the DM in of form of an ordinary differential equation (ODE), and improve sampling efficiency via utilizing faster ODE solver. Jolicoeur-Martineau \etal~\cite{jolicoeur2021gotta} introduce an advanced SDE solver to accelerate the reverse process via an adaptively larger sampling rate. 
Bao \etal~\cite{bao2022analytic} estimate the variance and KL divergence using the Monte Carlo method and a pretrained score-based model with derived analytic forms, which are simplified from the score-function.
In addition to those training-free methods, Luhman \& Luhman~\cite{luhman2021knowledge} compress the reverse denoising process into a single-step model; San-Roman~\cite{san2021noise} dynamically adjust the trajectory during inference. Nevertheless, implementing those methods requires additional training after obtaining a pretrained DM, which makes them less desirable in most situations. In summary, all those DM acceleration methods can be categorized into finding effective sampling trajectories. 

However, we show in this paper that, in addition to finding short sampling trajectories, diffusion models can be further \textit{accelerated through network compression for each noise estimation iteration}. Note that our method \ptq~is an orthogonal path with those above-mentioned fast sampling methods, which means it can be deployed as a plug-and-play module for those methods. To the best of our knowledge, our work is the first study on quantizing diffusion models in a post-training manner. 

\subsection{Post-training Quantization}

Quantization is one of the most effective ways to compress a neural network.
There are two types of quantization methods: Quantization-aware training (QAT) and Post-training quantization (PTQ). 
QAT~\cite{jacob2018qat,esser2020lsq,shang2022lipschitz,shang2022network} considers the quantization in the network training phase.
While PTQ~\cite{nagel2021whitepaper} quantizes the network after training.
As PTQ consumes much less time and computation resources, it is widely used in network deployment.

Most of the work of PTQ is to set the quantization parameters for weights and activcations in each layer.
Take uniform quantization as an example, the quantization parameters include scaling factor $s$ and zero point $z$.
A floating-point value $x$ is quantized to integer value $x_int$ according to the parameters:
\begin{equation}
    x_{int}=\text{clamp}(\lfloor \frac{x}{s} \rceil-z,p_{min},p_{max}).
\end{equation}
The clamp function clip the rounded value $\lfloor \frac{x}{s} \rceil-z$ to the range of $[p_{min},p_{max}]$.
In order to set quantization parameters for the weight tensor and the activation tensor in a layer, a simple but effective way is to select the quantization parameters that minimize the MSE of the tensors before and after quantization~\cite{banner2019ptq4bit,chokroun2019lowbitquant,wu2020easyquant,hubara2020layterwisecalibration}.
Other metrics, such as L1 distance, cosine distance, and KL divergence, can also be used to evaluate the distance of the tensors before and after quantization~\cite{migacz20178bittensorrt,yuan2022ptq4vit}.

In order to calculate the activations in the network, a small number of calibration samples should be used as input in PTQ.
The selected quantization parameters are dependent with the selection of these calibration samples.
\cite{hubara2020layterwisecalibration,li2021brecq,nahshan2021lossaware} demonstrate the effect of the number of the calibration samples.
Zero-shot quantization (ZSQ)~\cite{cai2020zeroq,he2021generative,zhong2022intraq} is a special case of PTQ.
ZSQ generates the calibration dataset according to information recorded in the network, such as the mean and var in batch normalization layer.
They generate the input sample by gradient descent method to make the distribution of the activations in network similar to the distribution of real samples.
The image generation process from noise in diffusion model only uses the network inference, which is quite different from previous ZSQ methods.

\section{PTQ on Diffusion Models}
\subsection{Preliminaries}
\label{sec:preliminary}

\begin{figure*}[tbh]
\centering
\includegraphics[width=1\textwidth]{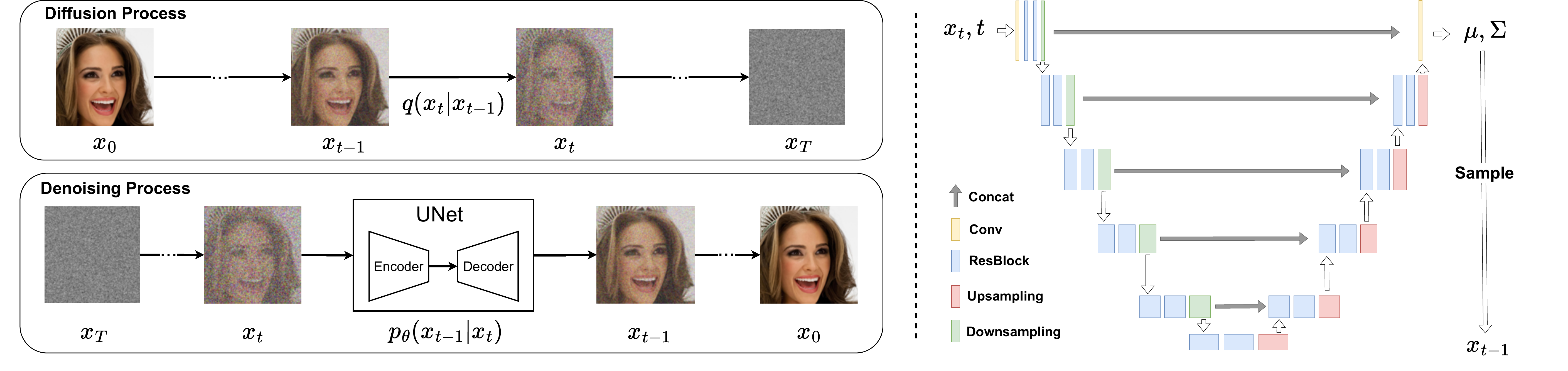}
      \caption{Brief illustration of Diffusion Model. The inference of diffusion models is extremely slow due to their two fundamental characteristics: \textbf{(1, Left)} the lengthy iterative process for denoising from noise input to synthetic images; and \textbf{(2, Right)} the cumbersome networks for estimating the noise in each denoising iteration.} 
  \vspace{-0.4cm}
\label{fig:diffusion_illustration}
\end{figure*}

\noindent\textbf{Diffusion Models.} The diffusion probabilistic model (DPM) is initially introduced by Sohl-Dickstein et al.~\cite{sohl2015deep}, where the DPM is trained by optimizing the variational bound $L_\text{VLB}$. Here, we briefly review the diffusion model to illustrate the difference from traditional models. Here, we briefly review the diffusion model, especially its lengthy diffusion and denoising process. We highlight that those properties make it difficult to simply generalize common PTQ methods into diffusion models simply in Sec.~\ref{sec:challenge}. 

Given a real data distribution $x_{0}\sim q(x_{0})$, we define the \textit{diffusion process} that gradually adds a small amount of isotropic Gaussian noise with a variance schedule $\beta_{1},...,\beta_{T}\in (0, 1)$ to produce a sequence of latent $x_{1},...,x_{T}$, which is fixed to a Markov chain. When $T$ is sufficiently large $T\sim\infty$ and a well-behaved schedule of $\beta_{t}$, $x_{T}$ is equivalent to an isotropic Gaussian distribution. 
\begin{align}
    q(\mathbf{x}_t \vert \mathbf{x}_{t-1}) &= \mathcal{N}(\mathbf{x}_t; \sqrt{1 - \beta_t} \mathbf{x}_{t-1}, \beta_t\mathbf{I}) \\
    q(\mathbf{x}_{1:T} \vert \mathbf{x}_0) &= \prod^T_{t=1} q(\mathbf{x}_t \vert \mathbf{x}_{t-1})
\label{equa:q_sampling}
\end{align}
A notable~\cite{ho2020denoising} property of the \textit{diffusion process} admits us to sample $x_{t}$ at an arbitrary timestep $t$ via directly conditioned on the input $x_{0}$. Let $\alpha_t = 1 - \beta_t$ and $\bar{\alpha}_t = \prod_{i=1}^T \alpha_i$:
\begin{equation}
    q(\mathbf{x}_t \vert \mathbf{x}_0) = \mathcal{N}(\mathbf{x}_t; \sqrt{\bar{\alpha}_t} \mathbf{x}_0, (1 - \bar{\alpha}_t)\mathbf{I})
\label{equa:q_xt_x0}
\end{equation}
Since $q(x_{t-1}|x_{t})$ depends on the data distribution $q(x_{0})$, which is intractable. Therefore, we need to parameterize a neural network to approximate it:
\begin{equation}
p_\theta(\mathbf{x}_{t-1} \vert \mathbf{x}_t) = \mathcal{N}(\mathbf{x}_{t-1}; \boldsymbol{\mu}_\theta(\mathbf{x}_t, t), \boldsymbol{\Sigma}_\theta(\mathbf{x}_t, t))
\label{equa:p_theta}
\end{equation}

We utilize the variational lower bound to optimize the negative log-likelihood. $L_\text{VLB} =$
\begin{equation}
\mathbb{E}_{q(\mathbf{x}_{0:T})} \Big[ \log \frac{q(\mathbf{x}_{1:T}\vert\mathbf{x}_0)}{p_\theta(\mathbf{x}_{0:T})} \Big] \geq - \mathbb{E}_{q(\mathbf{x}_0)} \log p_\theta(\mathbf{x}_0)
\label{equa:e_vlb}
\end{equation}
The objective function of the variational lower bound can be further rewritten to be a combination of several KL-divergence and entropy terms (more details in~\cite{sohl2015deep}).
\begin{align}
L_\text{VLB} &= \mathbb{E}_q [\underbrace{D_\text{KL}(q(\mathbf{x}_T \vert \mathbf{x}_0) \parallel p_\theta(\mathbf{x}_T))}_{L_T} \underbrace{- \log p_\theta(\mathbf{x}_0 \vert \mathbf{x}_1)}_{L_0} \nonumber \\
& + \sum_{t=2}^T \underbrace{D_\text{KL}(q(\mathbf{x}_{t-1} \vert \mathbf{x}_t, \mathbf{x}_0) \parallel p_\theta(\mathbf{x}_{t-1} \vert\mathbf{x}_t))}_{L_{t-1}} ]
\label{equa:vlb}
\end{align}
$L_{0}$ uses a separate discrete decoder derived from $\mathcal{N}(\mathbf{x}_{0}; \boldsymbol{\mu}_{\theta}(\mathbf{x}_1, 1), \boldsymbol{\Sigma}_{\theta}(\mathbf{x}_1, 1))$. $L_{T}$ does not depend on $\theta$, it is close to zero if $q(x_{T}|x_{0})\approx \mathcal{N}(0,I)$. The remain term $L_{t-1}$ is a KL-divergence to directly compare $p_{\theta}(x_{t-1}|x_{t})$ to \textit{diffusion process} posterior that is tractable when $x_{0}$ is conditioned,
\begin{align}
    \tilde{\beta}_t &:= \frac{1 - \bar{\alpha}_{t-1}}{1 - \bar{\alpha}_t} \cdot \beta_t \\ \tilde{\boldsymbol{\mu}}_t (\mathbf{x}_t\mathbf{x}_0) &:= \frac{\sqrt{\alpha_t}(1 - \bar{\alpha}_{t-1})}{1 - \bar{\alpha}_t} \mathbf{x}_t + \frac{\sqrt{\bar{\alpha}_{t-1}}\beta_t}{1 - \bar{\alpha}_t} \mathbf{x}_0 \\
    q(\mathbf{x}_{t-1} \vert \mathbf{x}_t, \mathbf{x}_0) &:= \mathcal{N}(\mathbf{x}_{t-1}; \tilde{\boldsymbol{\mu}}(\mathbf{x}_t, \mathbf{x}_0), \tilde{\beta}_t \mathbf{I}).
\label{equa:q_xt-1_xt_x0}
\end{align}
This is the training process of the diffusion model. 
After obtaining the well-trained noise estimation model $p_\theta(\mathbf{x}_{t-1} \vert\mathbf{x}_t)$ in Eq.~\ref{equa:p_theta}, given a random noise, we can generate samples through the \textit{denoising process} by iterative sampling $\mathbf{x}_{t-1}$ from $p_\theta(\mathbf{x}_{t-1} \vert\mathbf{x}_t)$ until we receive $\mathbf{x}_0$. 
Detailed information can be found in the surveys~\cite{yang2022diffusionsurvey,croitoru2022diffusionsurvey}. 
Since the iterative process for denoising from noise input to synthetic images is extremely long (\eg, pioneer work, DDPM~\cite{ho2020denoising} requires 4000 steps for generating a sample from noise), as illustrated in Fig.~\ref{fig:diffusion_illustration} (left); and the networks for estimating the noise in each denoising iteration is very deep and complicated, as illustrated in Fig.~\ref{fig:diffusion_illustration} (right). The inference of the diffusion model is expensive.


\noindent\textbf{Post-training Quantization} takes a well-trained network and selects the quantization parameters for the weight tensor and activation tensor in each layer.
We use the quantization parameters, scaling factor $s$, and zero point $z$ to transform a tensor to the quantized tensor\footnote{We focus on uniform quantization since it is the most widely used.}.
One of the most widely used methods to select the parameters is to minimize the error caused by quantization.
The quantization error $L_{quant}$ is formulated as:
\begin{equation}
    X_{sim} =s(\text{clamp}(\lfloor \frac{X_{fp}}{s} \rceil-z,p_{min},p_{max})+z),
\end{equation}
\begin{equation}
    L_{quant} =\text{Metric}(X_{sim},X_{fp}),
\end{equation}
where $X_{sim}$ is the de-quantized tensor, and $\text{Metric}$ is the metric function to evaluate the distance of $X_{sim}$ and the full-precision tensor $X_{fp}$.
MSE, cosine distance, L1 distance, and KL divergence are commonly used metric functions.
The quantization process can be formulated as: 
\begin{equation}
    \arg\min_{s,z}L_{quant}.
\label{equ:quantization_object}
\end{equation}

\begin{figure*}[ht]
\centering
\includegraphics[width=0.93\textwidth]{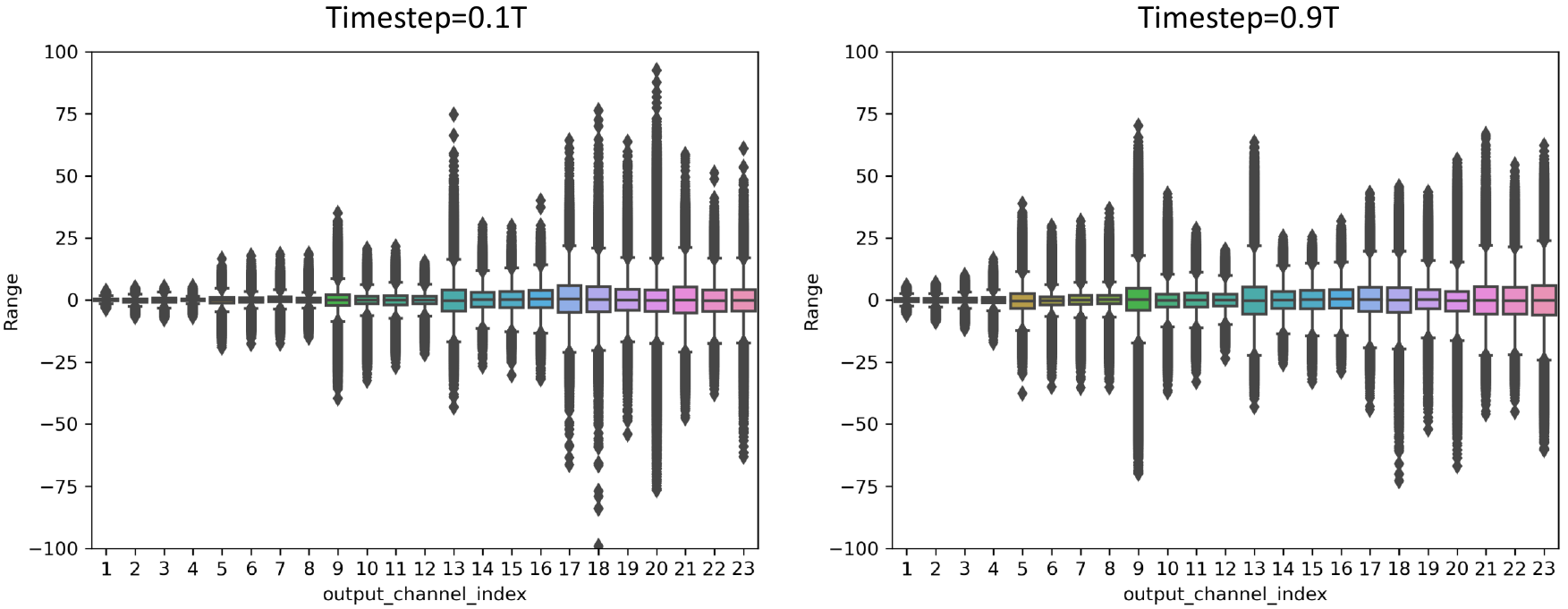}
\includegraphics[width=0.93\textwidth]{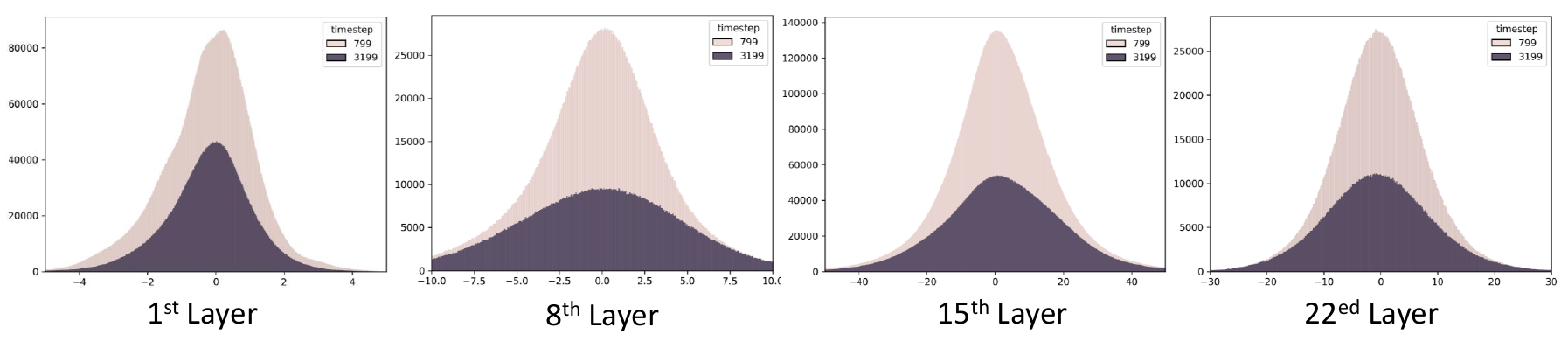}
\vspace{-0.1in}
\caption{Studies on the activation distribution \wrt time-step. \textbf{(Upper)} Per (output) channel weight ranges of the first depthwise-separable layer in diffusion model on different timestep. In the boxplot, the min and max values, the 2nd and 3rd quartile, and the median are plotted for each channel. We only include the layers in the decoder of UNet for noise estimation, as the ranges of the encoder and decoder are quite different. \textbf{(Bottom)} Histograms of activations on different time-steps by various layers. We can observe that the distribution of activations changes dramatically with time-step, which makes traditional single-time-step PTQ calibration methods inapplicable for diffusion models.} 
\vspace{-0.2in}
\label{fig:activation_distribution}
\end{figure*}

We can directly quantize the weight to minimize the quantization error, but we cannot get the activation tensor and quantize it without input.
In order to collect the full-precision activation tensor, a number of unlabeled input samples (calibration dataset) are used as input.
The size of the calibration dataset (\eg, 128 randomly selected images) is much smaller than the training dataset.

In general, PTQ quantizes a network in three steps:
\textbf{(i)} Select which operations in the network should be quantized and leave the other operations in full-precision. For example, some special functions such as softmax and GeLU often takes full-precision~\cite{shen2020qert}.Quantizing these operations will significantly increase the quantization error and they are not very computationally intensive; \textbf{(ii)} Collect the calibration samples. The distribution of the calibration samples should be as close as possible to the distribution of the real data to avoid over-fitting of quantization parameters on calibration samples; \textbf{(iii)} Use the proper method to select quantization parameters for weight tensors and activation tensors.

In the next sections, we will explore how to apply PTQ to the diffusion model step by step.

\subsection{Exploration on Operation Selection}
\label{sec:challenge}

For the diffusion model, we will analyze the image generation process to determine which operations should be quantized.
The diffusion model iteratively generate the $\mathbf{x}_{t-1}$ from $\mathbf{x}_{t}$.
At each timestep, the inputs of the network are $\mathbf{x}_{t}$ and $t$, and the outputs are the mean $\boldsymbol{\mu}$ and variance $\boldsymbol{\Sigma}$.
Then $\mathbf{x}_{t-1}$ is sampled from the distribution defined as Eq~\ref{equa:p_theta}.
As shown in Figure~\ref{fig:diffusion_illustration}, the network in the diffusion model often takes UNet-like CNN architecture.
The same as most previous PTQ methods, the computation-intensive convolution layers and fully-connected layers in the network should be quantized.
The batch normalization can be folded into the convolution layer.
The special functions such as SiLU and softmax are kept in full-precision.

There are two more questions for the diffusion model:
1. whether the network's outputs, $\boldsymbol{\mu}$ and $\boldsymbol{\Sigma}$, can be quantized?
2. whether the sampled image $\mathbf{x}_{t-1}$ can be quantized?
To answer the two questions, we only quantize the operation generating $\boldsymbol{\mu}$, $\boldsymbol{\Sigma}$, or $\mathbf{x}_{t-1}$.
As shown in Table~\ref{tab:opeartion_selection}, we observe that they are not sensitive to quantization and we indicate that they can be quantized.

\begin{table}[tb]
\centering
\caption{Exploration on operation selection for 8-bit quantization. The diffusion model is for unconditional ImageNet 64x64 image generation with a cosine noise schedule. DDIM (250 timesteps) is used to generate 10K images. IS is the inception score.}
\begin{tabular}{@{}cccc@{}}
\toprule
                  & IS    & FID   & sFID  \\ \midrule
FP                & 14.88 & 21.63 & 17.66 \\
quantize $\mu$    & 15.51 & 21.38 & 17.41 \\
quantize $\Sigma$ & 15.47 & 21.96 & 17.62 \\
quantize $x_{t-1}$    & 15.26 & 21.94 & 17.67 \\
quantize $\mu$+$\Sigma$+$x_{t-1}$      & 14.94 & 21.99 & 17.84 \\ \bottomrule
\end{tabular}
\label{tab:opeartion_selection}
\vspace{-0.2in}
\end{table}

\subsection{Exploration on Calibration Dataset}
\label{sec:calibration_dataset}

The second step is to collect the calibration samples for quantizing diffusion models.
The calibration samples can be collected from the training dataset for quantizing other networks.
However, the training dataset in the diffusion model is $\mathbf{x}_0$, which is not the network's input.
The real input is the generated samples $\mathbf{x}_{t}$.
Should we use the generated samples in \textit{diffusion process} or the generated samples in \textit{denoising process}?
At what time-step $t$, should the generated samples be collected?
This section will explore how to make a good calibration dataset. 

By all-inclusively investigating several intuitive PTQ baselines, we obtain four meaningful observations (Sec.~\ref{sec:exploration_calib}), which accordingly guide the design of our method (Sec.~\ref{sec:exploration_method}). Experimental results demonstrate that our method is efficient and effective. Through devised \ptq~calibration, the 8-bit post-training quantized diffusion model can perform at the same performance level as its full-precision counterpart, \eg, 8-bit diffusion model reaches 23.9 FID and 15.8 IS, while 32-bit one has 21.6 FID and 14.9 IS.

\subsubsection{Analysis on PTQ Calibration and DMs}
\label{sec:exploration_calib}

As discussed in Sec.~\ref{sec:challenge}, we desire the distribution of the collected calibration samples should be as close as possible to the distribution of the real data. In this way, the calibration set can supervise the quantization by minimizing the quantization error. Since previous works are implemented on single-time-step scenarios (\eg, CNN and ViT for image recognition and object detection)~\cite{nagel2021whitepaper,li2021brecq}, they can directly collect samples from the real training dataset for quantizing networks. Due to the small size of the calibration dataset, its collection is extremely sensitive. If the distribution of the collected dataset is not representative of the real dataset, it can easily lead to overfitting for the calibration task.

We encounter more challenges when calibrating PTQ for DM. Since the inputs of the to-be-quantized network are the generated samples $\mathbf{x}_{t}~(t=0,1,\cdots,T)$, in which $T$ is a large number to maintain the diffusion process converging to isotropic Normal distribution. 
To quantize the diffusion model, we are required to design a novel and effective calibration dataset collection method in this particular multi-time-step scenario.
We start by investigating both PTQ calibration and DMs, and then obtain the following instructive observations. 

\noindent\textbf{Observation 0: Distributions of activations changes along with time-step changing.}

To understand the output distribution change of diffusion models, we investigate the activation distribution with respect to time-step. We would like to analyze the output distribution at different time-step, for example, given $t_1=0.1T$ and $t_2=0.9T$, the output activation distributions of $p_\theta(\mathbf{x}_{{t_1}-1} \vert\mathbf{x}_{t_1})$ and $p_\theta(\mathbf{x}_{{t_2}-1} \vert\mathbf{x}_{t_2})$. Theoretically, if the distribution changes \wrt time-step, it would be difficult to implement previous PTQ calibration methods, as they are proposed for temporally-invariant calibration~\cite{nagel2021whitepaper,li2021brecq}. We first analyze the overall activation distributions of the noise estimation network via boxplot as~\cite{nagel2021whitepaper} did, and then we take a closer look at the layer-wise distributions via histogram. The results are shown in Fig.~\ref{fig:activation_distribution}. 
We can \textbf{observe} that at different time-steps, the corresponding activation distributions have large discrepancies, which makes previous PTQ calibration methods~\cite{nagel2021whitepaper,li2021brecq} inapplicable for multi-time-step models (\ie, diffusion models).

\begin{figure}[tb]
    \centering
    \includegraphics[width=0.4\textwidth]{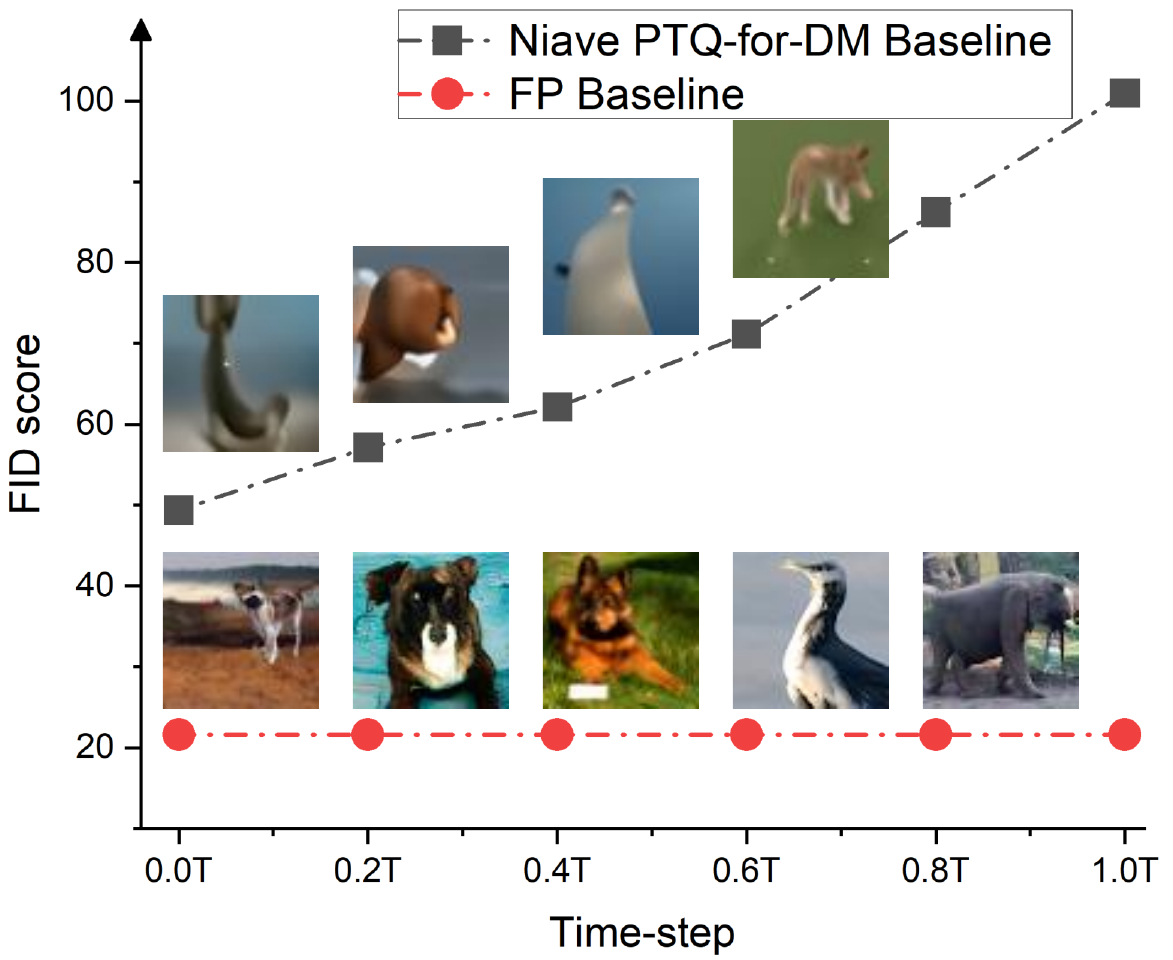}
    \vspace{-0.1in}
    \caption{Analyses of this calibration baseline at different time-steps. FP Baseline denotes the 32-bit model, which does not require to be calibrated.}
    \label{fig:timestep}
    \vspace{-0.1in}
\end{figure}

\noindent\textbf{Observation 1: Generated samples in the denoising process are more constructive for calibration.}

In general, there are two directions to generate samples for PTQ calibration in diffusion: raw images as input for diffusion process, and noise as input for denoising process. 
Previous PTQ methods use raw images, as raw images can serve as ground truth, representing the training set's distribution. We conduct a pair of comparison experiments, in which we separately collect two calibration sets with raw images for diffusion process and Gaussian noise for denoising process, and use these two sets to calibrate quantized models. 
Another similar intuitive baseline is to use the training samples in the diffusion process as calibration data. Specifically, we randomly generate a timestep $t$ for each image $\mathbf{x}_0$, and use Eq.~\ref{equa:q_xt_x0} according to $t$ to generate $\mathbf{x}_t$. In other word, collect calibration samples in a ``Image + Gaussian Noise'' manner. We name this scheme as \textbf{training-mimic} baseline. 
The results are listed in Tab.~\ref{tab:observation_1}.
We find that the input noises for diffusion process are more constructive for calibrating quantized DMs. 

\begin{table}[tb]
\centering
\caption{Results of calibration using noise (input of the denoising process), image (input of the diffusion process), and samples generated by Eq.~\ref{equa:q_sampling} (Mimicking the diffusion model training).}
\scalebox{0.85}{
\begin{tabular}{@{}cccc@{}}
\toprule
                  & IS $\uparrow$   & FID $\downarrow$  & sFID $\downarrow$ \\ \midrule
Noise Samples  &  13.92 &  33.15  &  20.38  \\
Image Samples    & 6.90 &  128.63 &  90.04 \\
Training-mimic    & 12.91 &  34.55 &  25.18 \\
\bottomrule
\end{tabular}}
\label{tab:observation_1}
\vspace{-0.2in}
\end{table}

\noindent\textbf{Observation 2: Sample $\mathbf{x}_t$ close to real image $\mathbf{x}_0$ is more beneficial for calibration.}

Based on the aforementioned observations, we establish a baseline of PTQ calibration for DM based on~\cite{nahshan2021lossaware}, in which the quantized diffusion models are calibrated with samples at time-step $t$, \ie, a set of $\mathbf{x}_t$. We refer to this straight-forward approach as a naive PTQ-for-DM baseline. 
Specifically, given a set of Gaussion noise $\mathbf{x}_T\sim \mathcal{N}(\mathbf{0},\mathbf{I})$, we use the diffusion model with the full-precision noise estimation network, $p_\theta(\mathbf{x}_{t-1} \vert\mathbf{x}_t)$ in Eq.~\ref{equa:p_theta} to generate a set of $\mathbf{x}_t$ as calibration set. Then as described in Sec.~\ref{sec:preliminary}, we use this collected set to calibrate our quantized noise estimation network, $p_{\theta^{\prime}}(\mathbf{x}_{t-1} \vert\mathbf{x}_t)$, in which $\theta^{\prime}$ is the quantized parameters. 
We conduct a series of experiments with this calibration baseline in different time-steps, \ie, $t=0.0T,0.2T,\cdots,1.0T$, where $T$ is the total denoising time-steps. The results are presented in Fig.~\ref{fig:timestep}. We can see that the 8-bit model calibrated by this naive baseline cannot synthesize satisfying images quantitatively and qualitatively. 

Fortunately, there is a windfall from these experiments. The PTQ calibration helps more when the time-step $t$ approaches the real image $\mathbf{x}_0$. 
There is an intuitive explanation for this observation. In the denoising process, with $t$ decreasing, the distribution of outputs of network $p_\theta(\mathbf{x}_{t-1} \vert\mathbf{x}_t)$ is similar to real images' distribution, which is a more significant phase in the image generation process.


\noindent\textbf{Observation 3: Instead of a set of samples generated at the same time-step, calibration samples should be generated with varying time-steps.}

Since our calibration dataset is collected for a multi-time-step scenario, while the common methods are proposed for single-time-step scenarios. We hypothesize that the calibration dataset for diffusion models should contain the samples with various time-steps, \ie, the calibration set should reflect the discrepancy of sample \wrt time-step. 
A straightforward way to test this hypothesis is to generate a set of uniformly sampled $t$ over the range of time-steps, \ie,
\begin{equation}
    t_{i}\sim U(0,T)~~(i=1,2,\cdots,N),
\end{equation}
where $U(0,T)$ is a uniform distribution between $0$ and $T$, $N$ is the size of calibration set, and $T$ is the number of time-steps in denoising process. Then given a Gaussion noise $\mathbf{x}_T\sim \mathcal{N}(\mathbf{0},\mathbf{I})$ and $t_i$, we utilize the diffusion model with the full-precision noise estimation network, $p_\theta(\mathbf{x}_{t-1} \vert\mathbf{x}_t)$ in Eq.~\ref{equa:p_theta} to generate a $\mathbf{x}_{t_i}$. Finally, we get the calibration set, $\mathcal{C}=\{\mathbf{x}_{t_i}\}_{i=1}^{N}$. Calibration samples can thus cover a wide range of time steps. We testify the effectiveness of this collection method, and present the results in Tab.~\ref{tab:observation_3}. The result validates our hypothesis that calibration samples should reflect the time-step discrepancy.

\begin{table}[tb]
\centering
\caption{Quantitative results of the intuitive baselines for the observations and our proposed \emph{NDTC} calibration method. With our method, the performance of PTQ for DM has been significantly improved, even exceeding full-precision DM performance \wrt IS and sFID.}
\scalebox{0.9}{
\begin{tabular}{@{}cccc@{}}
\toprule
                  & IS $\uparrow$   & FID $\downarrow$  & sFID $\downarrow$ \\ \midrule
Full precision DDIM        & 14.88 & 21.63 & 17.66 \\ \hline
Baseline in Observation 2    & 11.92 &  49.37  &  41.33  \\
Baseline in Observation 3  & 14.99 &  26.19 &  19.51 \\
\emph{NDTC (ours)} & \textbf{15.68} & \textbf{24.26} & \textbf{17.28}\\
\bottomrule
\end{tabular}}
\label{tab:observation_3}
\vspace{-0.2in}
\end{table}

\begin{figure}[!ht]  
    \centering
    \includegraphics[width=0.45\textwidth]{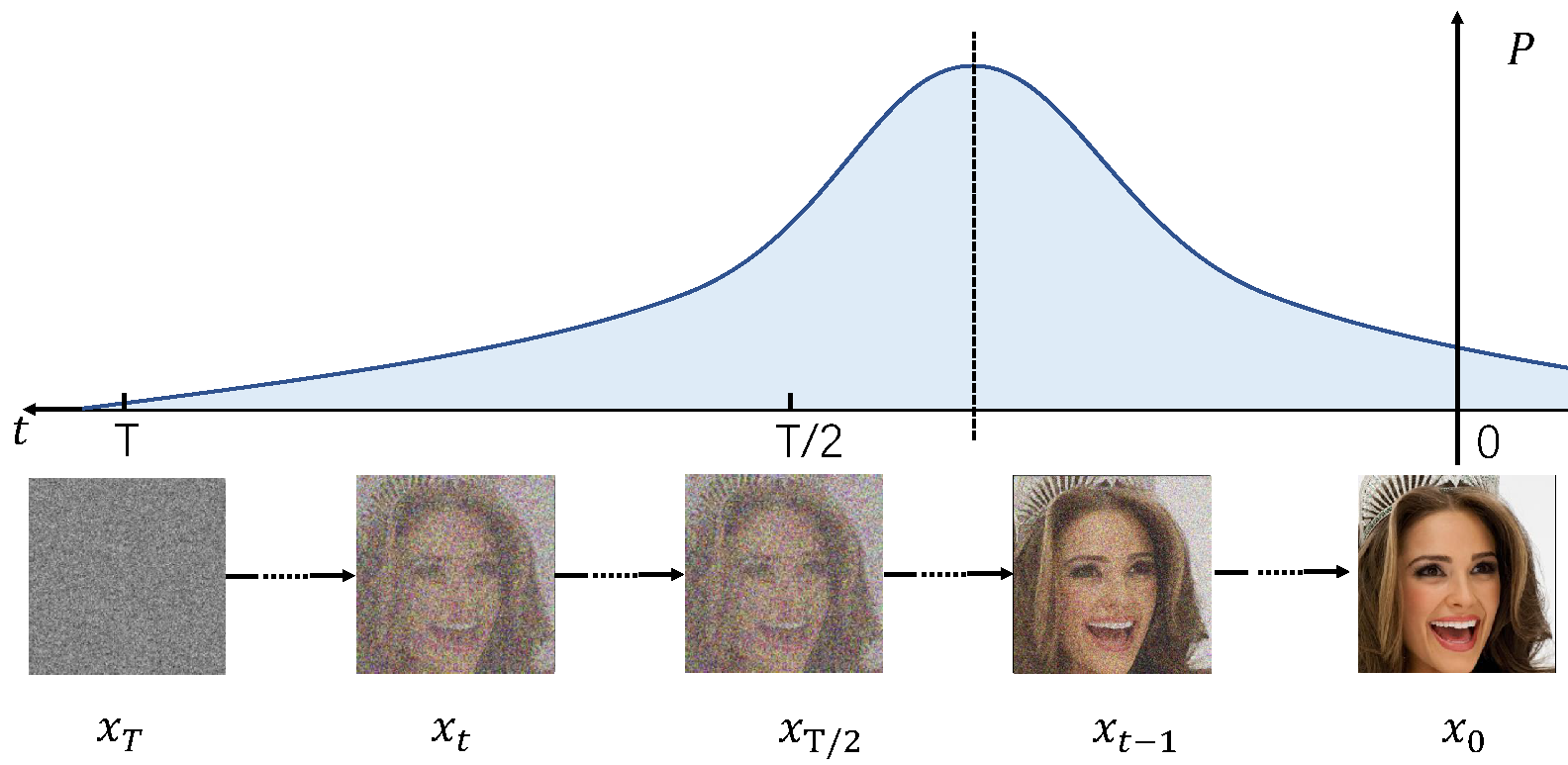}
    \caption{A general illustration of sampling time-steps following a distribution over the range of the denoising time-step.}
    \label{fig:sampling_t}
\end{figure}

\begin{algorithm}[!ht]  
\caption{Normally Distributed Time-step Calibration Collection (\emph{DNTC}) Algorithm.}  
\label{alg:NDTC}
\begin{algorithmic}[1] 
	\Require The size of calibration set $N$, and a mean of the Normal distribution $\mu$, and the full-precision noise estimation network $p_\theta(\mathbf{x}_{t-1} \vert\mathbf{x}_t)$ in Eq.~\ref{equa:p_theta}.
	\Ensure Obtain a Calibration Set $\mathcal{C}$.
	\State \textbf{Collecting Calibration Set}:
    \For{$i = 1$ to $N$}
    \State Sample $t_i$ from distribution $\mathcal{N}(\mu,\frac{T}{2})$ in Eq.~\ref{equ:timestep_normal};
    	\State Round down $t_i$ into a integer, \ie, $t_i = \lfloor t_i \rfloor$;
    	\State Clamp $t_i$ between $\left[0,T\right]$, \ie, $t_i = \text{Clamp}(0,T,t_i)$;
    	\State Produce sample on $t_i$ time-step:	
    	    \For{$t = T$ to $t_i$}
    	    \State Generate a Gaussian Noise $\mathbf{x}_T$ as initialization;
    	    \State Sample $\mathbf{x}_{t-1}$ using $p_\theta(\mathbf{x}_{t-1} \vert\mathbf{x}_t)$;
    	    \EndFor
    \State Output sample $x_{t_i}$;
    \EndFor
    \State Output a calibration set $\mathcal{C}=\{\mathbf{x}_{t_i}\}_{i=1}^{N}$.
\end{algorithmic}  
\end{algorithm}
\vspace{-0.2in}

\subsubsection{Normally Distributed Time-step Calibration}
\label{sec:exploration_method}

Based on the above-demonstrated calibration baselines and observations, we desire the calibration samples: \textbf{(1)} generated by the denoising process (from noise $\mathbf{x}_T$) with the full-precision diffusion model; \textbf{(2)} relatively close to $\mathbf{x}_0$, far away from $\mathbf{x}_T$; \textbf{(3)} covered by various time-steps. Note that (2) and (3) are a pair of trade-off conditions, which can not be satisfied simultaneously. 

Considering all the conditions, we propose a DM-specific calibration set collection method, termed as \textbf{N}ormally \textbf{D}istributed \textbf{T}ime-step \textbf{C}alibration (\textbf{\emph{NDTC}}). In this method, the calibration set $\{x_{t_i}\}$ are generated by the denoising process (for condition 1), where time-step $t_i$ are sampled from a skew Normal distribution (for balancing conditions 2 \& 3).
Specifically, we first generate a set of sampled $\{t_i\}$ following skew normal distribution over the time-step range (satisfying condition 3), \ie,
\begin{equation}
    t_{i}\sim \mathcal{N}(\mu,\frac{T}{2})~~(i=1,2,\cdots,N),
\label{equ:timestep_normal}
\end{equation}
where $\mathcal{N}(\mu,\frac{T}{2})$ is a normal distribution with mean $\mu\leq \frac{T}{2}$ and standard deviation $\sqrt{\frac{T}{2}}$, $N$ is the size of calibration set, and $T$ is the number of time-steps in denoising process. As $\mu$ is less than or equal to the median of time-step, $\frac{T}{2}$ (satisfying condition 2). Then given a Gaussian noise $\mathbf{x}_T\sim \mathcal{N}(\mathbf{0},\mathbf{I})$ and $t_i$, we utilize the diffusion model with the full-precision noise estimation network, $p_\theta(\mathbf{x}_{t-1} \vert\mathbf{x}_t)$ in Eq.~\ref{equa:p_theta} to generate a $\mathbf{x}_{t_i}$ (satisfying condition 1). The above-mentioned process of sampling time-steps is presented in Fig.~\ref{fig:sampling_t}. Finally, we get the calibration set, $\mathcal{C}=\{\mathbf{x}_{t_i}\}_{i=1}^{N}$. The detailed collection algorithm is presented in Alg.~\ref{alg:NDTC}.

\begin{figure}[t!]
\centering
\includegraphics[width=0.47\textwidth]{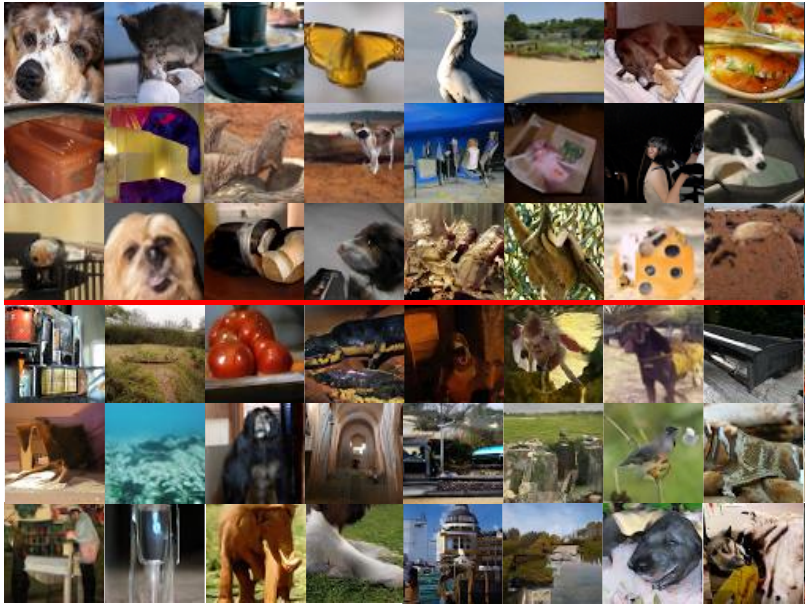}
\caption{Non-cherry-picked generated samples. \textbf{(Upper)} Samples synthesized by full precision DDPM~\cite{ho2020denoising}. \textbf{(Bottom)} Samples synthesized by 8-bit model quantized by our method. Note that \ptq~can directly output an 8-bit diffusion with the pre-trained 32-bit diffusion model as input in a \textit{training-free} manner.} 
\label{fig:demo}
\end{figure}

The effectiveness of \emph{NDTC} is assessed by comparing it to the mentioned PTQ baselines and full-precision DMs. The results are presented in Tab.~\ref{tab:observation_3} and Fig.~\ref{fig:demo}. 

\subsection{Exploration on Parameter Calibration}

When the calibration samples are collected, the third step is selecting quantization parameters for tensors in the diffusion model.
In this section, we explore the metric to calibrate the tensors.
As shown in Table~\ref{tab:calibration_metric}, the MSE is better than the L1 distance, cosine distance, and KL divergence.
Therefore, we take MSE as the metric for quantizing the diffusion model.

\begin{table}[tb]
\centering
\caption{Exploration on calibration metric for 8-bit quantization. We set p=2.4 for MSE metrics.}
\scalebox{1.0}{
\begin{tabular}{@{}cccc@{}}
\toprule
               & IS $\uparrow$   & FID $\downarrow$  & sFID $\downarrow$ \\ \midrule
L1 distance     & 7.38            & 100.52 & 63.01 \\
Cosine distance & 12.85           & 34.81  & 23.75 \\
KL divergence   & 11.74           & 47.27  & 45.08 \\
MSE             & 13.76           & 30.46  & 19.42 \\ \bottomrule
\end{tabular}}
\label{tab:calibration_metric}
\end{table}

\section{More Experiments}

We select the diffusion models that generating CIFAR10~\cite{krizhevsky2009learningCIFAR} $32\times 32$ images or ImageNet~\cite{deng2009imagenet} down-sampled $64\times 64$ images.
We experiment on both DDPM (4000 steps) and DDIM (100 and 250 steps) to generate the images.

We use the proposed method in Section~\ref{sec:calibration_dataset} to generate 1024 calibration samples.
And we quantize the network to 8-bit.
Then we sample 10,000 images for evaluation.
The results are listed in Table~\ref{tab:more_experiments}.
Note that the number of samples that we generate is only 10,000 (50,000 in several papers) in order to efficiently compare our methods with other baselines quantitatively. Thus some reported results in this paper are slightly different from the results in the original papers. 
There is an exciting result in these experiments. In the setting of using DDPM~\cite{ho2020denoising} to generate images with the size of $32\times32$, the 8-bit DDPM quantized by our method outperforms the full-precision DDPM. 
As discussed in Sec.~\ref{sec:intro}, there are two factors slowing down the denoising process: i) lengthy iterations for sampling images from noise, and ii) a cumbersome network for estimating noise in each iteration. 
The successes of previous DM acceleration methods~\cite{chen2020wavegrad,san2021noise,nichol2021improved,song2020score,bao2022analytic,lu2022dpm} validate the existence of model redundancy from the perspective of iteration length. With this exciting result, we uncover the redundancy from a previously unknown perspective, in which the noise estimation network is also redundant.

\begin{table}[tbh]
\centering
\caption{Experiment on 8-bit quantized diffusion models generating CIFAR10 image or ImageNet image.}
\begin{tabular}{@{}c|cccc@{}}
\toprule
Task            & Method & IS $\uparrow$   & FID $\downarrow$  & sFID $\downarrow$ \\ \midrule
ImageNet 64x64  & FP     & 15.38 & 21.70 & 17.93  \\
DDIM 100 steps  & \ptq     & 15.52 & 24.92 & 17.36 \\ \midrule
ImageNet 64x64  & FP     & 14.88 & 21.63 & 17.66 \\
DDIM 250 steps  & \ptq     & 15.88 & 23.96 & 17.67 \\ \midrule
ImageNet 64x64  & FP     & 15.93 & 20.82 & 17.42 \\
DDPM 4000 steps & \ptq    & 15.28 & 23.64 & 17.29 \\\midrule
CIFAR 32x32     & FP                  & 9.18  & 10.05 & 19.71 \\
DDIM 100 steps  & \ptq & 9.31  & 14.18 & 22.59 \\ \midrule
CIFAR 32x32     & FP                  & 9.19  & 8.91  & 18.43 \\
DDIM 250 steps  & \ptq & 9.70  & 11.66 & 19.71 \\ \midrule
CIFAR 32x32     & FP                  & 9.28  & 7.14  & 17.09 \\
DDPM 4000 steps & \ptq & 9.55  & 7.10  & 17.02 \\ \midrule
\end{tabular}
\label{tab:more_experiments}
\end{table}

\section{Conclusion}
\label{sec:con}
Two orthogonal factors slow down the denoising process: i) lengthy iterations for sampling images from noise, and ii) a cumbersome network for estimating noise in each iteration. Different from mainstream DM acceleration works focusing on the former, our work digs into the latter. In this paper, we propose Post-Training Quantization for Diffusion Models (\ptq), in which a pre-trained diffusion model can be directly quantized into 8 bits without experiencing a significant degradation in performance. 
Importantly, our method can be added to other fast-sampling methods, such as DDIM~\cite{nichol2021improved}.


{\small
\bibliographystyle{ieee_fullname}
\bibliography{manuscript}
}

\clearpage
\section{Appendix}
\subsection{Nonuniform Distribution Selection}
Notably, we have carefully chosen the nonuniform distribution for sampling timestep $t$ (Eq.~\ref{equ:timestep_normal}). Specifically, except for the normal distribution in the paper, we also consider the Poisson and exponential distributions. Also, a series of hyperparameter selection experiments are conducted. More details are presented in Tab.~\ref{table:nonuniform-distributions}.

\begin{table}[!ht]
    \centering
    \caption{Hyperparameter selection for non-uniform distribution in Algorithm~\ref{alg:NDTC}.}
    \scalebox{0.8}{
\begin{tabular}{@{}c|cccc@{}}
\toprule
Task            & Method & IS $\uparrow$   & FID $\downarrow$  & sFID $\downarrow$ \\ \midrule
                                   & FP     & 14.88 & 21.63 & 17.66 \\ \cline{2-5}
Other Nonuniform                  & Poisson     & 13.29 & 34.54 & 25.84 \\ 
Distributions                     & Exponential & 12.87 & 39.91 & 30.04 \\ \hline
                                  & $\mu=\frac{T}{2}$,~ $\sigma=0.5\sqrt{\frac{T}{2}}$    &  15.45 &  25.11 &  17.35 \\
Normal Distribution               & $\mu=\frac{T}{2}$,~ $\sigma=1.0\sqrt{\frac{T}{2}}$    & 15.65 & 24.83 & 18.90 \\
   with Different                 & $\mu=\frac{T}{2}$,~ $\sigma=2.0\sqrt{\frac{T}{2}}$    & 15.85 & 24.27 & 17.92 \\ \cline{2-5}
      Mean $\mu$,                 & $\mu=\frac{1.5T}{2}$,~ $\sigma=\sqrt{\frac{T}{2}}$    & 12.63 &  39.09 & 35.81 \\
and Variance $\sigma$             & $\mu=\frac{1.0T}{2}$,~ $\sigma=\sqrt{\frac{T}{2}}$   & 15.65 & 24.83 & 18.90 \\
                                  & $\mu=\frac{0.5T}{2}$,~ $\sigma=\sqrt{\frac{T}{2}}$    & 15.88 & 23.96 & 17.67 \\ \midrule

\end{tabular}}
\label{table:nonuniform-distributions}
\end{table}

\subsection{Actual Acceleration}
We test the latency(ms) of the original network (provided checkpoint) and the quantized network on Nvidia RTX A6000 GPU. The results in Table~\ref{tab:actual_speedup} show that the 8-bit quantization achieves about 2x speedup.
The speedup can be more significant on NPU.
\begin{table}[h!]
    \centering
\caption{Inference speed test with Nvidia RTX A6000.}
\scalebox{1.0}{
\begin{tabular}{@{}cccc@{}}
\toprule
Task                                                                      & Batch Size & FP32  & INT8  \\ \midrule
\multirow{2}{*}{\begin{tabular}[c]{@{}c@{}}ImageNet\\ 64x64\end{tabular}} & 1          & 9.80  & 4.99  \\
                                                                          & 16         & 64.42 & 28.16 \\ \midrule
\multirow{2}{*}{\begin{tabular}[c]{@{}c@{}}CIFAR\\ 32x32\end{tabular}}    & 1          & 5.92  & 2.98  \\
                                                                          & 16         & 23.15 & 14.13 \\ \bottomrule
\end{tabular}}
\label{tab:actual_speedup}
\end{table}

\end{document}